\documentclass[journal]{IEEEtran}
\usepackage{amsmath,amsfonts}
\usepackage{algorithmic}
\usepackage{algorithm}
\usepackage{array}
\usepackage[caption=false,font=normalsize,labelfont=sf,textfont=sf]{subfig}
\usepackage{textcomp}
\usepackage{stfloats}
\usepackage{setspace}
\usepackage{url}
\usepackage{verbatim}
\usepackage{graphicx}
\usepackage{cite}
\usepackage{multirow}
\usepackage{booktabs}
\usepackage{color}
\usepackage{threeparttable}
\usepackage{array}
\usepackage{diagbox}
\usepackage{graphics}
\usepackage{graphicx}
\usepackage{epsfig}
\usepackage{epstopdf}
\usepackage{hyperref}
\usepackage{bbding}
\usepackage[numbers,sort&compress]{natbib}

\begin{document}

\title{GMFL-Net: A Global Multi-geometric Feature Learning Network for Repetitive Action Counting}

\author{Jun Li,~\IEEEmembership{Senior Member,~IEEE}, Jinying Wu,~\IEEEmembership{}Qiming Li$^{*}$,~\IEEEmembership{Senior Member,~IEEE}, and Feifei Guo~\IEEEmembership{}

\thanks{\hspace{-1mm}*Corresponding author.}}

%


\maketitle

\begin{abstract}
With the continuous development of deep learning, the field of repetitive action counting is gradually gaining notice from many researchers. Extraction of pose keypoints using human pose estimation networks is proven to be an effective pose-level method. However, existing pose-level methods suffer from the shortcomings that the single coordinate is not stable enough to handle action distortions due to changes in camera viewpoints, thus failing to accurately identify salient poses, and is vulnerable to misdetection during the transition from the exception to the actual action. To overcome these problems, we propose a simple but efficient Global Multi-geometric Feature Learning Network (GMFL-Net). Specifically, we design a MIA-Module that aims to improve information representation by fusing multi-geometric features, and learning the semantic similarity among the input multi-geometric features. Then, to improve the feature representation from a global perspective, we also design a GBFL-Module that enhances the inter-dependencies between point-wise and channel-wise elements and combines them with the rich local information generated by the MIA-Module to synthesise a comprehensive and most representative global feature representation. In addition, considering the insufficient existing dataset, we collect a new dataset called Countix-Fitness-pose (\href{https://github.com/Wantong66/Countix-Fitness}{https://github.com/Wantong66/Countix-Fitness}) which contains different cycle lengths and exceptions, a test set with longer duration, and annotate it with fine-grained annotations at the pose-level. We also add two new action classes, namely lunge and rope push-down. Finally, extensive experiments on the challenging RepCount-pose, UCFRep-pose, and Countix-Fitness-pose benchmarks show that our proposed GMFL-Net achieves state-of-the-art performance.

\end{abstract}

\begin{IEEEkeywords}
Repetitive action counting, multi-geometric information, global feature learning.
\end{IEEEkeywords}

\section{Introduction}
\IEEEPARstart{R}{epetitive} action counting is an essential task in computer vision for analysing various human activities. For example, it is widely used in many fields such as sports training \cite{fitness1,fitness2}, intelligent surveillance \cite{is1,is2}, video understanding \cite{video1,video2}, and three-dimensional reconstruction \cite{humanaction1, humanaction2}. However, as a relatively new topic in the research community, repetitive action counting has not been thoroughly studied. It thus still remains a challenging problem due to difficulties in key information extraction, under-utilisation of geometric information between joints and so on.

\begin{figure*}[t]  
    \centering
    \includegraphics[width=18cm]{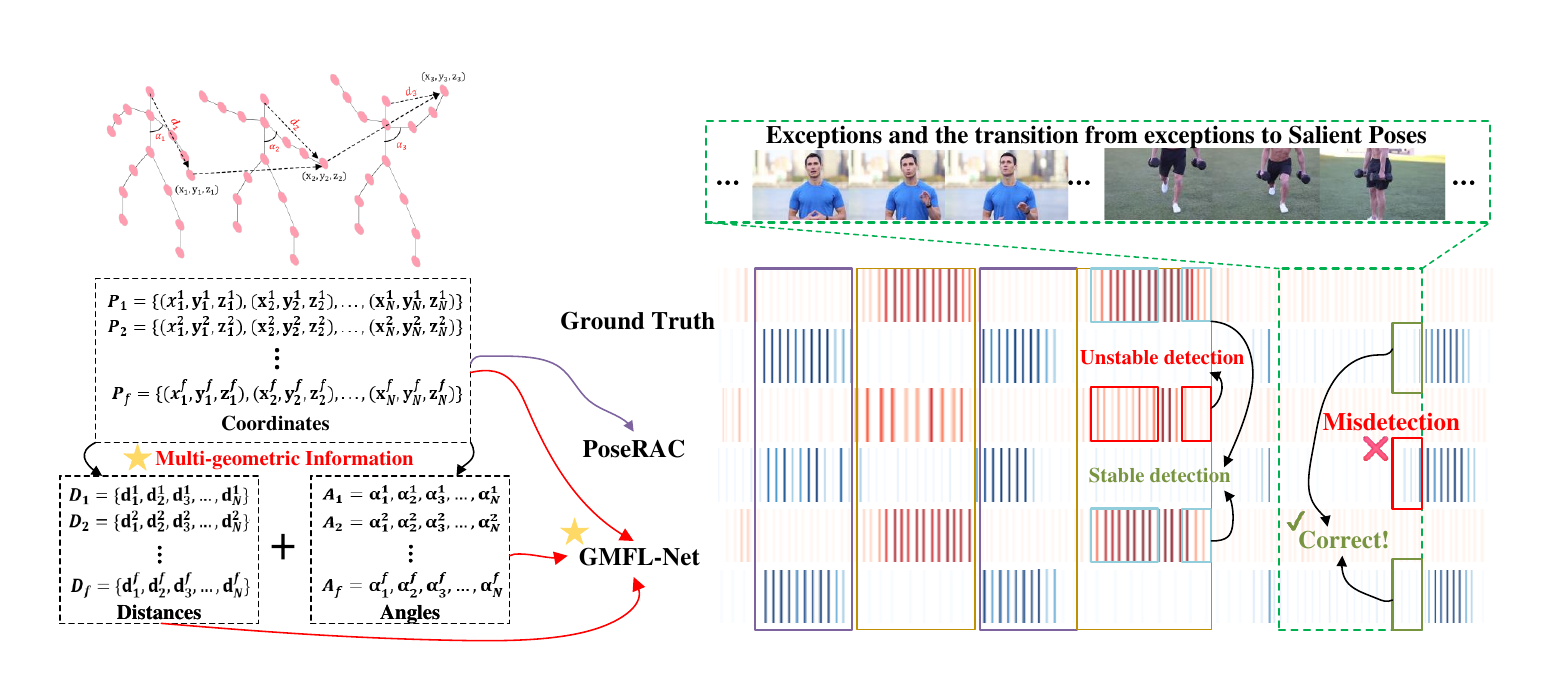}
    \caption{Coordinates $P_i$ ($i=1,2,...f$) contains the coordinates of the $N$ joints in each frame $f$. Distance $D_i$ ($i=1,2,...f$) contains the distance between every two joints in each frame $f$. Angle $A_i$ ($i=1,2,...f$) contains the angle between every three joints in each frame $f$. To keep the number of features consistent with the joint coordinates, we randomly selected $N$ distance and angle features. A darker red color in the graph means it is more likely to represent salient pose I, while a darker blue color means it is more likely to represent salient pose II.}
    \label{fig1}
\end{figure*}

The dominant repetitive action counting methods in recent years can be categorized into two major classes: video-level and pose-level methods. In video-level methods, exemplified by the works of Levy \emph{et al.} \cite{levy} and Pogalin \emph{et al.} \cite{pogalin}, a common practice is to assume the periodicity of the action usually fixed at a predefined time scale. Nevertheless, different repetitive actions usually exhibit varying cycle lengths, the predefined timescales could potentially affect the accuracy of the count. Therefore, the video-level method TransRAC \cite{hu} performs multiscale processing on the input video sequences and achieves state-of-the-art performance by using density map prediction as the cycle predictor and introducing fine-grained action cycle annotations into the dataset. However, the video-level methods \cite{dwibedi,x3d,zhang,liu,swin,huang,hu,tcsvt} tend to involve significant redundant information, including background details, requiring expensive feature extraction and complex interactions with the video context.

In order to solve the problems of video-level methods, the pose-level methods \cite{poserac,chen} introduce the human pose estimation techniques \cite{pose3d,liu,song,su,zhangxikun,blazepose,fang,xu} into the task of repetitive action counting. Pose keypoints can be represented by using lightweight 3D coordinate positions of human joints, relative to the RGB video and depth data, which are better able to reflect the human movement. PoseRAC \cite{poserac} introduces a Pose Saliency Representation (PSR) mechanism that uses the two most salient poses to represent each action, providing a more efficient alternative to the video-level representation of RGB frames. However, current pose-level methods focus on salient poses using only joint coordinates. As shown in Fig.~\ref{fig1}, this single coordinate is not stable enough to handle action distortions caused by changes in camera viewpoints, thus failing to accurately identify salient poses, and is vulnerable to misdetection during the transition from the exception to the actual action. In addition, these pose-level methods that solely rely on Transformer for feature extraction and modelling of human joint coordinates focus on long-range dependencies but often overlook local detail features and motion trends, thus fail to take full advantage of the interrelationship between local and global features.

Inspired by \cite{chen,actionlet,lan}, we found that the skeleton of the body is closely related to action. The interrelationship of joints as connecting structures of skeleton is present in every action sequence. The primitive joints of the human body contain substantial hidden geometric information, such as angles and distances between joints. This geometric information of the changes in angles and distances between joints, which is inspired by the brain's perception of action recognition, could well represent the interactions between body parts and is crucial for accurate action recognition. Additionally, since different actions cause distinct changes in geometric information that are significant during movement, combining the multi-geometric information can improve the model's stability in handling viewpoint changes and exceptions, making it more efficient at recognizing repetitive actions. Therefore, this paper proposes a Global Multi-geometric Feature Learning Network (GMFL-Net) for repetitive action counting. Specifically, we first propose a Multi-Geometric Information Aggregation Module (MIA-Module) that improves information representation by fusing multi-geometric features, followed by semantic similarity among the input multi-geometric features. As shown in Fig.~\ref{fig1}, our method effectively stabilises the effects of camera viewpoint changes by introducing multi-geometric information to assist the original coordinate information, and reduces misdetections during the transition from exception to the actual action in videos, thus making the final detection results closer to the ground truth. Secondly, we design a  Global Bilinear Feature Learning Module (GBFL-Module) to enhance the point feature representation from a global perspective. This module improves the inter-dependencies between point-wise and channel-wise elements and combines the inter-dependencies with the rich geometric and local information generated by the MIA-Module to synthesise a comprehensive and representative feature representation. Finally, the mapping relationships between features and action classes are established through the Classification Head. These relationships are transformed into scores for different action classes, which are then passed to the Repetitive Counting Module (RC-Module) to complete the counting task.

In addition, due to the limited diversity of existing pose-level datasets, we collect a new dataset called Countix-Fitness-pose. This dataset contains 553 videos and approximately 7,593 fine-grained pose-level annotations. To increase the challenge of the dataset, we carefully update videos with some exceptions, a test set with longer duration, and two new action classes: lunge and rope push-down. Subsequently, to fully evaluate the robustness and effectiveness of our proposed GMFL-Net, we conduct extensive experiments on three challenging repetitive action counting datasets (\emph{i.e.}, RepCount-pose, UCFRep-pose, and CountixFitness-pose). The results show that our GMFL-Net achieves state-of-the-art performance.

In summary, our research contributions are threefold:
\begin{itemize}
	\item We propose an innovative network architecture, GMFL-Net, which focuses on the introduction of the MIA-Module and the GBFL-Module. Through these two modules, we are able to efficiently use multi-geometric information to improve and stabilise the recognition of salient poses, while combining point-wise and channel-wise elements for global feature learning.
	\item We introduce a new dataset, Countix-Fitness-pose, which contains 553 videos covering different cycle lengths and exceptions, a test set with longer duration, and providing approximately 7, 593 fine-grained annotations at the pose-level. We also add some new action classes which provide a richer resource and a higher level of challenge for the study of repetitive action counting.
	\item We conduct extensive experiments on three challenging benchmark datasets, namely RepCount-pose, UCFRep-pose, and Countix-Fitness-pose. The experimental results demonstrate the state-of-the-art performance of our proposed network on the repetitive action counting task.
\end{itemize}

The remainder of our paper is organized as follows. Sec.~\ref{sec2} provides an overview of the related work. Sec.~\ref{sec3} presents the details of our GMFL-Net. Sec.~\ref{sec4} presents extensive experiments to validate the impact of different component in GMFL-Net on performance and to verify the effectiveness and robustness of our method on three datasets. Sec.~\ref{sec5} gives the conclusion of this paper.

\section{Related Work} \label{sec2}
\subsection{Video-level Repetitive Action Counting} \label{sec2.1}
In early methods, researchers convert video features into one-dimensional signals and extracted the periodicity of repetitive actions by means of Fourier transform \cite{Cutler,pogalin,Cyclic}, peak detection\cite{peak}, and so on. However, these methods are limited to dealing with static conditions. Runia \emph{et al.} \cite{realworld} use continuous wavelet transform to process the optical flow features for non-static and non-smooth video conditions to more accurately estimate the repetitive actions. Then, with the explosion of deep learning, a lot of methods \cite{levy,pogalin,x3d,liu,swin,huang,tcsvt} have emerged recently. For example, Zhang \emph{et al.} \cite{zhang} propose an innovative two-way context-aware regression model, incorporating a dual-period estimation strategy to improve the precision of action cycle prediction. And they introduce the UCFRep dataset comprising 526 videos. Similarly, RepNet \cite{dwibedi} proposes to use a temporal self-similarity matrix for predicting the dynamic periods of repetitive actions in videos and simultaneously creates a dataset called Countix, which contains about 6,000 videos. Nevertheless, these previous methods primarily rely on coarse-grained annotations and lack the capability to handle videos of varying lengths. To address these problems, TransRAC \cite{hu} designs a multiscale temporal correlation encoder, which not only accommodates high and low frequency actions, but also adapts to video sequences of different lengths. Moreover, TransRAC introduces the RepCount dataset, including 1,451 videos and about 20,000 fine-grained annotations. In addition, Jacquelin \emph{et al.} \cite{Jacquelin} explore an unsupervised method suitable for repetitive counting and apply it to time series data. Furthermore, the effective combination of acoustic and visual features to improve the accuracy of repetitive action counting is first achieved by \cite{sound}, demonstrating that multi-modal method can overcome the shortcomings of visual data alone.

\subsection{Pose-level Repetitive Action Counting Based on Human Pose Estimation}
Convolutional Neural Networks (CNNs) are dominant in previous human pose estimation methods \cite{openpose,posenet,alphapose,blazepose,stgcn,csi,wang}. \cite{transpose,xu,actionlet,cli,mliu,ydu,zhou} becomes widespread among researchers with the emergence of Vision Transformer \cite{vit} in various visual tasks. Nevertheless, although significant progress has been made in the field of human pose estimation \cite{alphapose,blazepose}, its application in repetitive action counting remains limited. To address the limitations of video-level methods, including inefficient key feature extraction and the presence of large amounts of redundant information, PoseRAC \cite{poserac} integrates a lightweight human pose estimation network (Blazepose \cite{blazepose}) into the repetitive action counting task. Furthermore, PoseRAC \cite{poserac} introduces the Pose Saliency Representation (PSR) mechanism, which uses the two most salient poses to represent the action. It reduces the computational complexity associated with extracting high-level semantic information from intra-frame spatial and inter-frame temporal in traditional RGB frame-based methods. However, pose-level methods \cite{poserac,chen} ignore the importance of changes in multi-geometric information of human joints during motion. As a result, they fail to effectively deal with the effects of viewpoint changes and exceptions in videos. To address these issues, we introduce the MIA-Module, which leverages multi-geometric information (\emph{i.e.}, coordinates, angles, and distances between joints), to stabilise and enhance salient pose recognition by focusing on changes in feature details through local aggregation.

\subsection{Global Feature Learning}
Global feature learning \cite{stgcn,geo-lstm,large-kernel,Towards} plays a key role in deep learning as it enables the effective understanding and integration of context information. NLP \cite{attention,nlp} is highly dependent on the interaction of global contextual information, which is essential for improving the efficiency of the task. The attention mechanism \cite{attention} has emerged as a key driver for the rapid development of this field. In recent years, Convolutional Neural Network (CNN)-based methods achieve remarkable results in computer vision. However, existing CNN-based methods often fail to fully exploit the pixel-by-pixel global context information for modelling. In fact, the global spatio-temporal context can effectively eliminate local interference and thus improve the characterisation of target features. Therefore, methods such as ViT \cite{vit} and Swin Transformer \cite{swin} introduce Transformer and its variants into the image fields. Meanwhile, IIP-Transformer \cite{iip} and ST-TR \cite{sttr} also apply Transformer to human pose estimation, significantly improving network performance by integrating global context information. Currently, the attention mechanism \cite{attention} is widely regarded as one of the best methods to achieve global feature learning. For example, SENet \cite{senet} reweights the global feature map using point-wise or channel-wise attention to enhance feature saliency. PCT \cite{pct} proposes the offset attention mechanism to efficiently construct global feature maps by dealing with the offsets between attention features. Similarly, Point Transformer \cite{pointtransformer} leverages Transformer blocks to construct efficient global features based on point data, demonstrating strong performance in  point cloud classification and segmentation. However, the introduction of the attention mechanism significantly increases the computational cost. Therefore, this paper proposes the GBFL-Module to balance the high-quality global feature learning with computational efficiency. This module efficiently extracts global features by capturing and fusing point-wise and channel-wise features based on the learned multi-geometric information.

\begin{figure*}[htp]
	\centering
	\includegraphics[width=18cm]{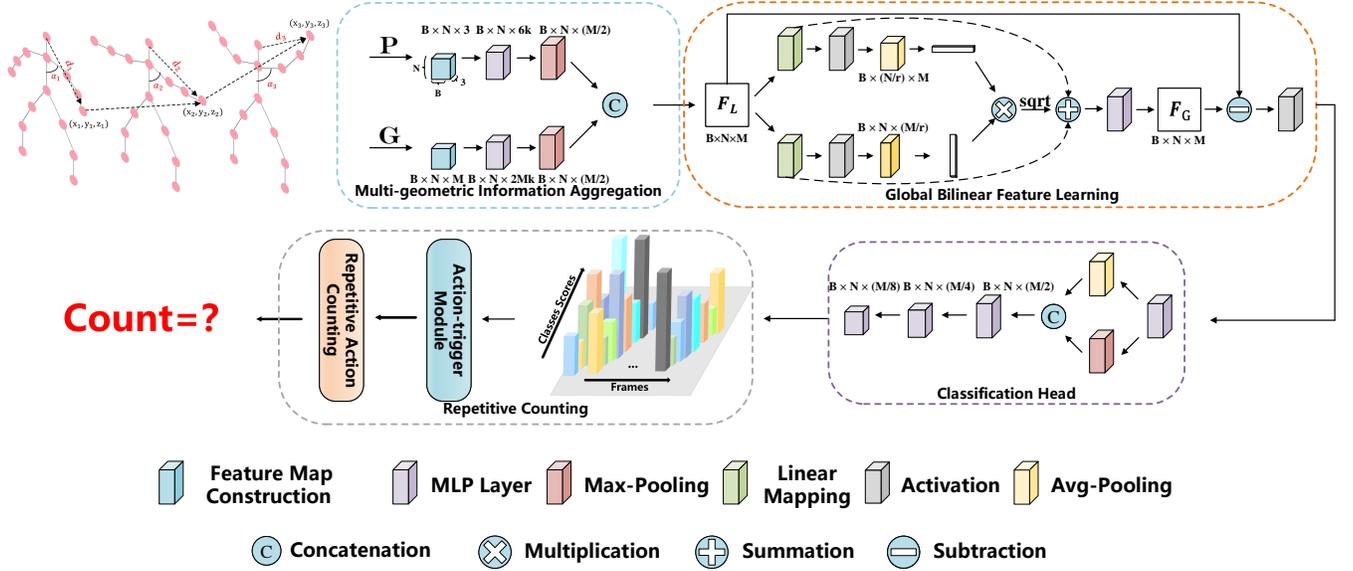}
	\caption{The overall architecture of GMFL-Net includes the MIA-Module, GBFL-Module, Classification Head, and RC-Module. The $(x, y, z)$ in the figure represent joint coordinates, $\alpha_1, \alpha_2, \alpha_3$ represent angles between joints, and $d_1, d_2, d_3$ represent distances between joints. $P$ represents the coordinate information and $G$ represents the rest of the geometric information, \emph{i.e.}, angles and distances between joints.}
	\label{fig2}
\end{figure*}

\section{THE PROPOSED METHOD} \label{sec3}
Given a video $V = \{f_i\}^{T} \in {R}^{C \times H \times W \times T}$ with $T$ frames, it is mapped between salient poses and action classes using the proposed network to obtain class scores $\hat{S}\in {R}^{C \times T}$ which are ultimately used to predict the number of repetitive actions $Y$ through the Action-trigger mechanism.
\subsection{Model Overview} \label{sec3.1}
\indent As shown in Fig.~\ref{fig2}, our method consists of three parts.
\begin{itemize}
	\item The first part is the MIA-Module (\S~\ref{sec3.2}). It first uses the offline and well-trained human pose estimation network BlazePose \cite{blazepose} to extract global joint coordinate information from each frame of a video sequence. Subsequently, action detail focus is enhanced by aggregating the feature mappings at each point using a k-Nearest Neighbours (kNN) method based on the semantic similarity between joints. Meanwhile, the stability of significant pose recognition is improved by fusing multiple geometric features captured through parallel branches using geometric information of different actions.
	\item The second part consists of the GBFL-Module and the Classification Head (\S~\ref{sec3.3}). The purpose of the GBFL-Module is to regulate the representation of features through the global learning of point-wise and channel-wise features. The Classification Head, on the other hand, is used to efficiently establish the mapping between pose and action classes. In addition, two loss functions (\S~\ref{sec3.4}) are used for training, namely Triplet Margin Loss and Binary Cross Entropy Loss.
	\item The third part is the RC-Module (\S~\ref{sec3.5}), which is used to count the number of repetitive actions when the salient action classification scores are obtained for all the frames of the entire video sequence.
\end{itemize}

\subsection{MIA-Module} \label{sec3.2}

In previous video-level methods (\emph{e.g.}, \cite{zhang,dwibedi,hu,i3d,c3d,p3d,x3d,swin,tsn}), the feature extraction stage involves a significant amount of redundant information, including background details and so on. To address this issue, we utilize the offline and well-trained human pose estimation network BlazePose \cite{blazepose} like in PoseRAC \cite{poserac} to extract the required global joint coordinate information. This improvement preserves the essential features, eliminates redundant information, and helps the network to focus on the core of the repetitive actions in the video.

\indent To begin with, we transform each frame in the video sequence into a sequence of joints using a human pose estimation network, which can be defined as:
\begin{equation}
	\begin{aligned}
		V &= \{f_i\}^{T} \in {R}^{C \times H \times W \times T}, \\
		V &\rightarrow P = \{p_i\}^{T} \in {R}^{N \times 3 \times T},
	\end{aligned}
	\label{eq1}
\end{equation}
where each $f_i$ represents each RGB frame in the video sequence, $C$ represents the number of channels, typically three RGB channels, $H$ represents the height, $W$ represents the width, and $T$ represents the number of frames. Each $p_i$ represents joints of each RGB frame, which is represented by the sequence $ N \times 3 \times T $. $3$ is the feature dimension of each joint, with two coordinates and one depth information respectively, and $N$ denotes the number of joint points.

To capture the local details of the action in motion, according to the k-NN($\mathcal{N}$)(n) algorithm, we can find the neighbours $\forall p_{\text{i}_k} \in \mathcal{N}(p_i)$ of a certain point $p_i$. By combining $p_i$ and its neighbours $\forall p_{\text{i}_k}$ under the 3D Euclidean distance metric, we can define a local feature map: $\hat{P_i}= [p_i, p_{\text{i}_k} - p_i]$, $\hat{P_i} \in R^{6k}$. Thus, the local feature maps $\hat{P}$ for all joints $P$ are defined as follows:
\begin{equation}
	\begin{aligned}
        \hat{P} = \{\hat{P}_1, \hat{P}_2, \ldots, \hat{P}_N\} \in R^{N \times 6k}.
	\end{aligned}
	\label{eq2}
\end{equation}

Subsequently, we encode the local feature maps $\hat{P}$ using an MLP $M_p$ and apply max-pooling over $k$ neighbours to aggregate the local context information:
\begin{equation}
	\begin{aligned}
        P = maxpooling(M_p(\hat{P})) \in R^{N \times M/2},
	\end{aligned}
	\label{eq3}
\end{equation}
where MLP contains a $1 \times 1$ convolution, a batchnorm, and an activation layer. $M$ is the output dimension of the convolution in the MLP.

Different actions have unique details that result in variations in joint angles and distances during movement. The hidden geometric information in these angles and distances helps distinguish among different action classes and plays a key role in recognizing repetitive actions. Therefore, we introduce the distance $d_{ab}^{i}$ and angle $\theta_{abc}^{i}$ information between joints as auxiliary geometric information $g_i$:

\begin{equation}
	\begin{aligned}
        & d_{ab} = \|\vec{p_a p_b}\|, \\
        &\theta_{abc} = \arccos(\frac{\vec{\mathit{p_a p_b}} \cdot \vec{p_b p_c}}{\|\vec{p_a p_b}\| \|\vec{p_b p_c}\|}), \\
        & g_i = \{d_{ab}^{i} , \theta_{abc}^{i} \},
	\end{aligned}
	\label{eq4}
\end{equation}
where $d_{ab}$, $\theta_{abc}$, and $g_i \in R^{V}$, $p_a, p_b, p_c$ are the three different joints in $p_i$. $V$ denotes the number of distances and angles.

Subsequently, we utilize two MLPs to map $G=\{g_1, g_2, \ldots, g_N\} \in {R}^{N \times V}$, which consists of all joints' auxiliary geometric information, into M-dimensional space $G \in {R}^{N \times M}$ for higher-dimensional features. For diverse feature learning from different perspectives, in parallel processing, we will similarly construct the local feature map of $\hat{G}_i = [g_i, g_{\text{i}_k} - g_i] \in R^{2Mk}$, where $\forall g_{\text{i}_k}$ are corresponding auxiliary geometric feature of $\forall p_{\text{i}_k} \in \mathcal{N}(p_i)$. Accordingly, the local feature map $\hat{G}$ of all auxiliary geometric features is represented as follows:
\begin{equation}
	\begin{aligned}
        \hat{G} = \{\hat{G}_1, \hat{G}_2, \ldots, \hat{G}_N\} \in R^{N \times 2Mk}.
	\end{aligned}
	\label{eq5}
\end{equation}

Following a procedure similar to that described in Equation (3), we can obtain a local feature map of the geometric information as follows:
\begin{equation}
	\begin{aligned}
        G = maxpooling(M_g(\hat{G})) \in R^{N \times M/2},
	\end{aligned}
	\label{eq6}
\end{equation}
where the MLP parameters of $M_g$ and $M_p$ are not shared with each other.

Finally, we concatenate the local context over $k$ neighbours $P$ and the geometric local features $G$ to obtain the local context of the multiple geometric information:
\begin{equation}
	\begin{aligned}
        F_L = Concat(P,G) \in R^{N \times M}.
	\end{aligned}
	\label{eq7}
\end{equation}

\subsection{GBFL-Module and Classification Head} \label{sec3.3}

In addition to capturing more local detailed features at geometric information, we also consider improving the overall feature mapping through global aggregation. It is well known that the attention mechanism is one of the most widely used modules to capture global dependencies, but it consumes a lot of memory and computational resources. Therefore, we employ an element-by-element global feature aggregation method based on point-wise and channel-wise to form global inter-dependencies between elements, thereby significantly reducing computational complexity.

First, we perform dimensionality reduction on the fused output features $F_L$ using $1 \times 1$ convolution $W_{conv} \in R^{N \times M/r}$ to reduce the complexity and number of parameters of the network, where $r$ is a reduction factor. Next, we apply the ReLU activation function to the features after the $1 \times 1$ convolution to introduce non-linearity. Finally, the values of $N$ joints within the local region are averaged by the avg-pooling operation along the point-wise dimension. This step aims to smooth the feature map while preserving the overall trend and important information of the features. The specific process is as follow:
\begin{equation}
    \begin{aligned}
        G_N = avgpooling(ReLU(W_{conv}(F_L))) \in R^{N/r \times M}.
    \end{aligned}
    \label{eq8}
\end{equation}

Similar to Equation (8), but the next operation is performed along the channel-wise dimension:
\begin{equation}
    \begin{aligned}
        G_C = avgpooling(ReLU(W_{conv}(F_L))) \in R^{N \times M/r}.
    \end{aligned}
    \label{eq9}
\end{equation}

Because $G_C$ captures the channel-wise dependencies and $G_N$ represents the context relationship between the whole joints, employing geometric means to compute the bilinear combination of these two features allows for a comprehensive synthesis and full retention of both types of global information, thus enhancing the feature representation. The specific processing is as follows:
\begin{equation}
    \begin{aligned}
        B = \sqrt{G_C \cdot G_N} ,
    \end{aligned}
    \label{eq10}
\end{equation}
where $B \in R^{N \times M/r}$ is the output of the global information aggregation.

To recover the channel dimensions and generate a global feature aggregation map, we use a MLP $M_{\phi}$ along with two residual connections.
\begin{equation}
    \begin{aligned}
        F_G = M_{\phi}(B + G_C + G_N) \in R^{N \times M}.
    \end{aligned}
    \label{eq11}
\end{equation}

\begin{figure}[h]  
	\centering
	\centering	
	\includegraphics[width=8cm,scale=1]{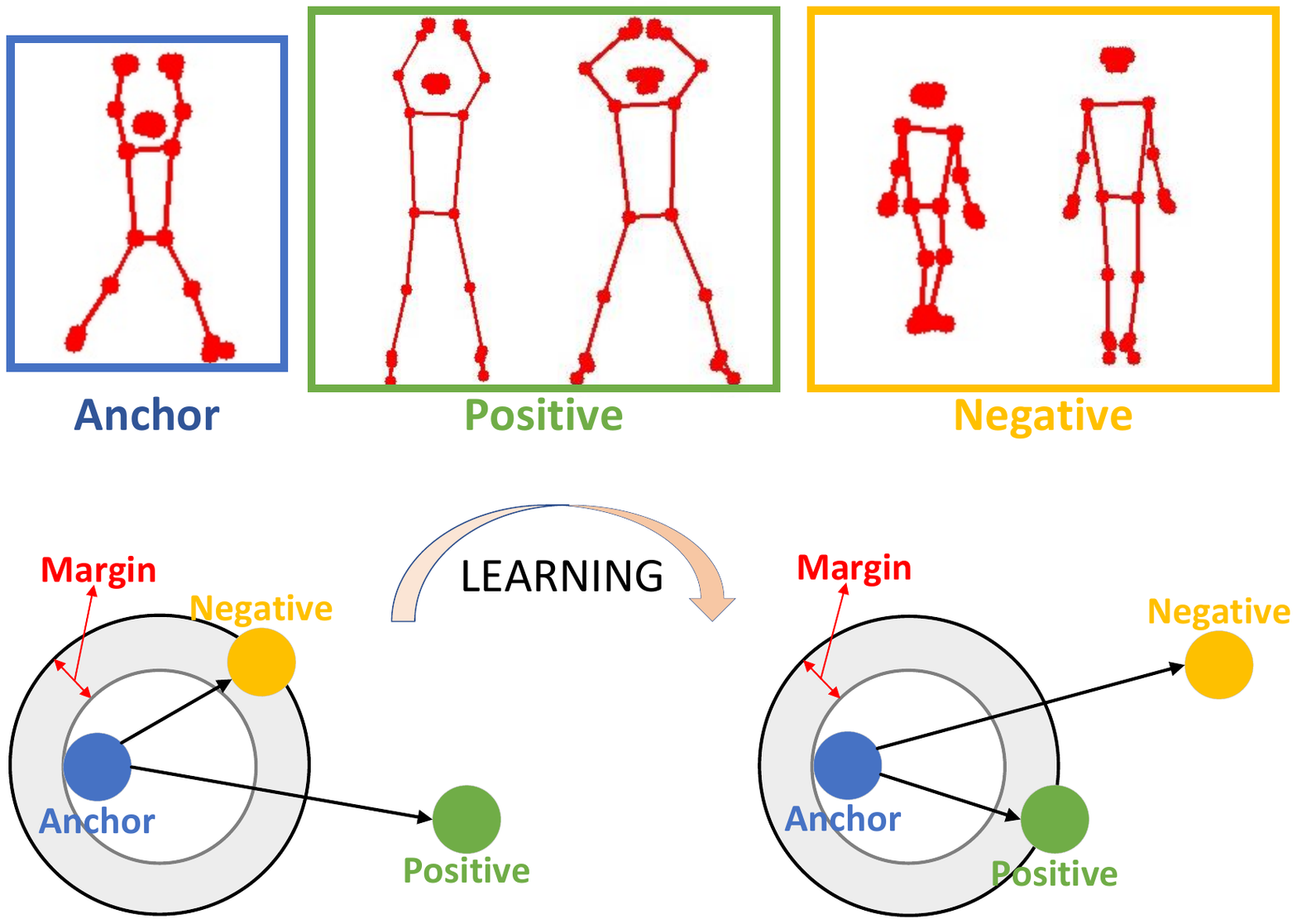}
	\caption{Illustration of Triplet Margin Loss. We use it to improve the Encoder. After training, the distance between the anchor and the positive example decreases, while the distance between the anchor and the negative example increases.}
	\label{fig3}
\end{figure}

\begin{figure*}[t]  
	\centering
	\centering	
	\includegraphics[width=18cm,scale=1]{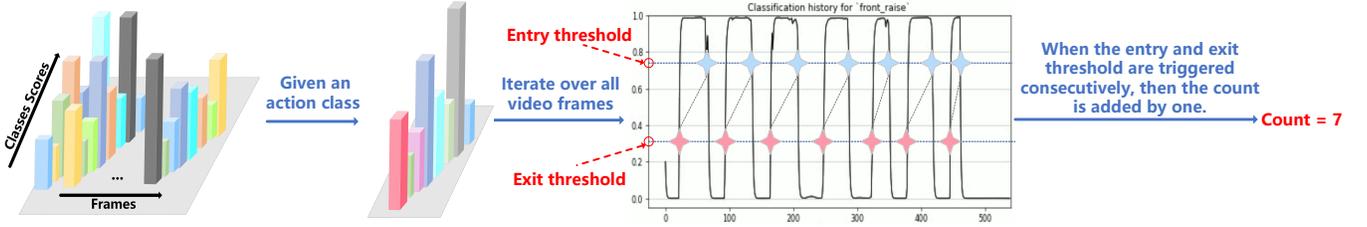}
	\caption{Illustration of the mechanism of RC-Module. We scan all frames and obtain scores $S_c$ for specific action class. In this process, we set entry thresholds and exit thresholds which are used to distinguish between two salient actions. When the score of Salient Pose I exceeds the entry threshold and the score of Salient Pose II is below the exit threshold, the mechanism of RC-Module triggers. The count is added one whenever Salient Poses I and II are triggered sequentially.}
	\label{fig4}
\end{figure*}

In addition, to reduce the negative impact of the absolute position of the local feature map $ F_L $ and to learn more representative and more salient features, we subtract the local feature $ F_L $ from its counterpart $ F_G $ to obtain the residual feature map. Next, an activation $\sigma$ is used to add more non-linearisation to the residual feature mapping.
\begin{equation}
    \begin{aligned}
        F_{\psi} = \sigma(F_G - F_L) \in R^{N \times M}.
    \end{aligned}
    \label{eq12}
\end{equation}

To predict the final classification scores $S\in {R}^O$ for all action classes in the current frame, we input $F_{\psi} \in R^{N \times M}$ into a feed-forward neural network consisting of a max-pooling, an avg-pooling, and three cascading MLP layers ($M_{\beta}$, $M_{\gamma}$, and $M_{\delta}$). The purpose of max-pooling and avg-pooling is to further refine the features of $F_{\psi}$. To reduce the number of parameters in the network, we used three MLPs with different parameter settings, each with output dimensions gradually reduced to 1024, 512, and 256, respectively. Finally, the pose mapping is completed by a linear layer to obtain the score $\hat{S}\in {R}^{O \times T}$ for a single frame, where $O$ denotes output channels. This process can be defined as:
\begin{equation}
	\begin{aligned}
        &F_{\eta} = Concat(maxpooling(F_{\psi}),avgpooling(F_{\psi})), \\
        &F_{out} = M_{\delta}(M_{\gamma}(M_{\beta}(F_{\eta}))), \\
		&\hat{S} = Linear(Flatten(F_{out})).
	\end{aligned}
	\label{eq13}
\end{equation}

\subsection{Losses and Metric Learning} \label{sec3.4}

In our module, we first introduce Metric Learning \cite{facenet,poserac}, namely the Triplet Margin Loss, to enhance the encoder. Through the training process, the encoder can learn higher-dimensional and more representative features $F_{out}$. Thereafter, anchors, positive samples of the same class, and negative samples of different classes in each batch are extracted and clustered in the high-level space using the Triplet Margin Loss:
\begin{equation}
	\begin{aligned}
		L_{\text{tri}} &= max(CS(a,p)-CS(a,n) + margin, 0),
	\end{aligned}
	\label{eq14}
\end{equation}
where $a$, $p$, $n$ denote anchors, positive samples, and negative samples. And $CS$ stands for cosine similarity, which is used to measure the similarity between features. As shown in Fig.~\ref{fig3}, our main goal is to decrease the feature distance between the anchors ($a$) and the positive sample ($p$) by optimizing Triplet Margin Loss, while simultaneously increasing the feature distance between the anchors ($a$) and the negative sample ($n$). By doing so, our network can achieve a better distinction between the poses of each.

Subsequently, to perform binary classification for each action class, we use the Binary Cross Entropy Loss for the classification:
\begin{equation}
	\begin{aligned}
		L_{\text{bce}} &= -\frac{1}{B}\sum_{i=1}^{B} (\frac{1}{C}\sum_{j=1}^{C}loss(i,j)), \\
		loss(i,j) &= y_{ij}\log(p_{ij}) + (1-y_{ij})\log(1-p_{ij}),
	\end{aligned}
	\label{eq15}
\end{equation}
where $B$ denotes the batch size, where each frame constitutes a batch, $C$ represents the number of classes, $y$ represents the true label, and $p$ is the result of our optimized prediction.

Finally, our training contains both types of losses:
\begin{equation}
	\begin{aligned}
		L_{\text{total}} &= L_{\text{bce}} + \alpha L_{\text{tri}},
	\end{aligned}
	\label{eq16}
\end{equation}
where $\alpha$ is a weighting factor controlling both losses, which ensures that the relative importance of the metric learning loss (Triplet Margin Loss) and the Binary Cross Entropy Loss are within the same range of values during network training.

\subsection{RC-Module} \label{sec3.5}
To obtain the final repetitive action count output $Y$ while keeping the network lightweight, we employ a streamlined Action-trigger Module. As shown in Fig.~\ref{fig4}, we first scan all frames of the input video and extract the action scores $\hat{S}\in {R}^{C \times T}$ from each frame. We then set upper and lower thresholds to distinguish the two salient poses, and add one to the count when the scores of the two salient poses exceed or fall below the upper and lower thresholds in turn. The upper and lower thresholds are determined by averaging the scores for Salient pose I and II.

\subsection{Implementation Details} \label{sec3.6}
\subsubsection{Training} \label{sec3.6.1}
During training, we utilize the RepCount-pose, UCFRep-pose, and Countix-Fitness-pose datasets with fine-grained pose-level annotations. Instead of using the entire video sequence as input, our training dataset includes only video frames with Salient pose I and Salient pose II, which helps improve the training speed and the network's fitting results.

\subsubsection{Inference} \label{sec3.6.2}
During inference, the entire video sequence is fed into our network. Each frame in the video is processed by the Encoder and Classification Head to obtain a score for each action class. These scores are then passed to the Action-trigger Module for repetitive action counting.

\section{Experiments} \label{sec4}

\begin{figure*}[t]  
	\centering
	\centering	
	\includegraphics[width=18cm]{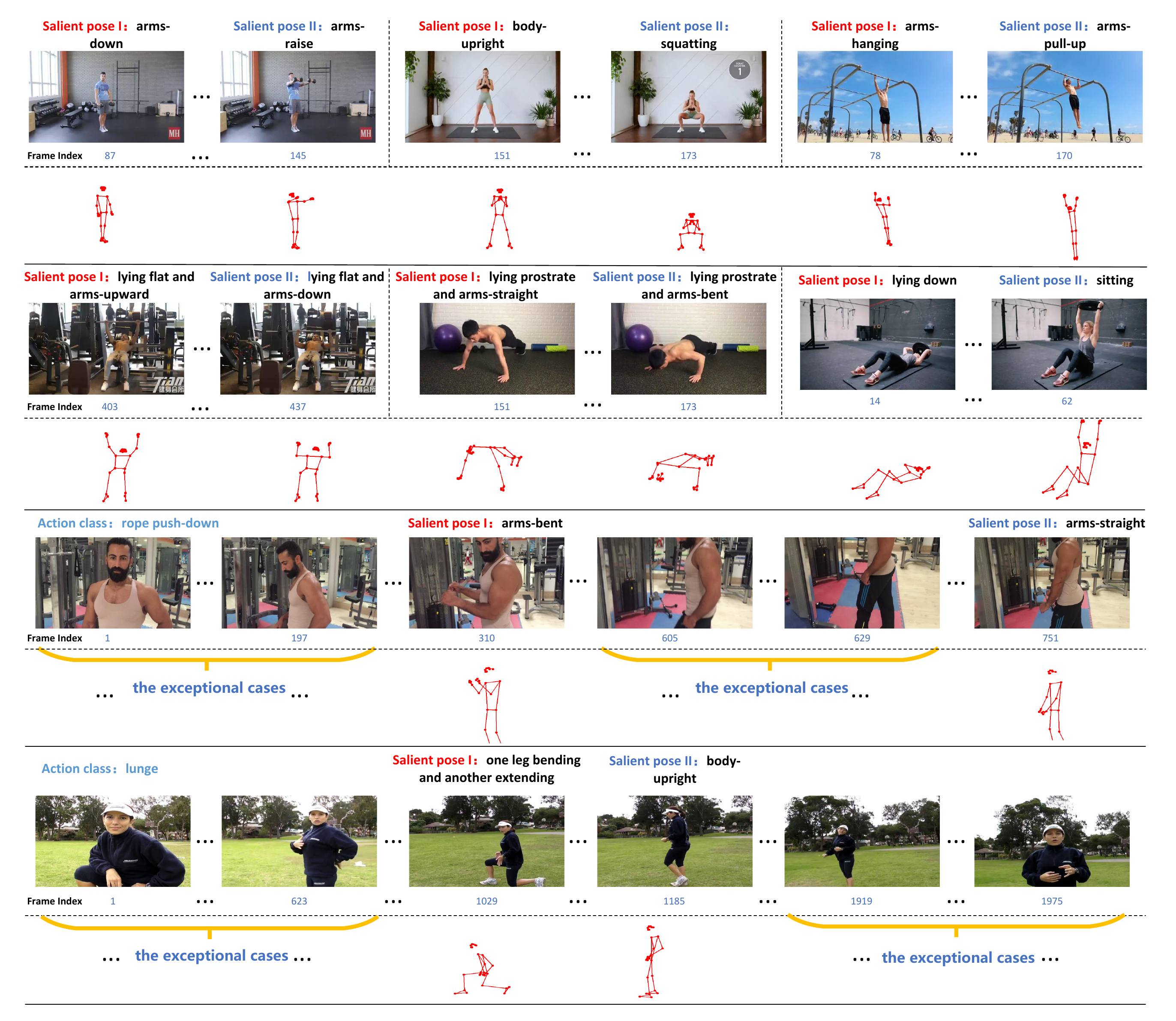}
	\caption{Illustration of the six action classes in our proposed dataset and the implementation of the PSR mechanism \cite{poserac}. We need to accurately select two salient poses that represent the completion of an action in the given videos, labelled as salient pose I and salient pose II. For example, at frame 87 of the given video, we select this frame as the representative of salient pose I, and at frame 145, we select this frame as the representative of salient pose II.}
	\label{fig5}
\end{figure*}

\begin{table*}[htp]
    \centering
    \caption{Comparisons between RepCount-pose, UCFRep-pose, and Countix-Fitness-pose,  including differences between action classes in the dataset as well as differences between salient poses for each action. RepCount, UCFRep, and Countix are RepCount-pose, UCFRep-pose and Countix-Fitness-pose respectively.}
    \label{table1}
    \begin{tabular}{@{}cccccc@{}}
        \toprule
        \multirow{2}{*}{Action class} & \multicolumn{3}{c}{Datasets} & \multicolumn{2}{c}{Salient poses} \\
        \cmidrule(lr){2-4} \cmidrule(lr){5-6}
        & RepCount & UCFRep & Countix & Salient pose I & Salient pose II \\
        \midrule
        bench-press & \Checkmark & \Checkmark & \Checkmark & lying flat and arms-upward & lying flat and arms-down \\
        front-raise & \Checkmark & \XSolidBrush & \Checkmark & arms-down & arms-raise \\
        push-up & \Checkmark & \Checkmark & \Checkmark & lying prostrate and arms-straight & lying prostrate and arms-bent \\
        pull-up & \Checkmark & \XSolidBrush & \Checkmark & arms-hanging & arms-pull-up \\
        sit-up & \Checkmark & \XSolidBrush & \Checkmark & lying down & sitting \\
        squat & \Checkmark & \Checkmark & \Checkmark & body-upright & squatting \\
        jumping jack & \Checkmark & \Checkmark & \XSolidBrush & body-upright and arms-down & jumping up and arms-upward \\
        pommel horse & \Checkmark & \Checkmark & \XSolidBrush & body leaning to the left & body leaning to the right \\
        lunge & \XSolidBrush & \XSolidBrush & \Checkmark & one leg bending and another extending & body-upright \\
        rope push-down & \XSolidBrush & \XSolidBrush & \Checkmark & arms-bent & arms-straight \\
        \bottomrule
    \end{tabular}
\end{table*}

\begin{table*}[htp]
	\centering
	\caption{Comparison between RepCount-pose, UCFRep-pose, and Countix-Fitness-pose, including the number of training set and test set videos, and event count of each action.}
	\label{table2}
	\begin{tabular}{ccccccccccrrrrrrrrr}
		\toprule
        & \multicolumn{4}{c}{RepCount-pose} & \multicolumn{4}{c}{UCFRep-pose} & \multicolumn{4}{c}{Countix-Fitness-pose}  \\
        \cmidrule(lr){2-5} \cmidrule(lr){6-9} \cmidrule(lr){10-13}
        \multirow{2}{*}{}{Action class} & \multicolumn{2}{c}{Training Set} & \multicolumn{2}{c}{Test Set} & \multicolumn{2}{c}{Training Set} & \multicolumn{2}{c}{Test Set} & \multicolumn{2}{c}{Training Set} & \multicolumn{2}{c}{Test Set} \\
		\cmidrule(lr){2-3} \cmidrule(lr){4-5} \cmidrule(lr){6-7} \cmidrule(lr){8-9} \cmidrule(lr){10-11}  \cmidrule(lr){12-13}
	    & Video & Event & Video & Event & Video & Event & Video & Event & Video & Event & Video & Event  \\
		\midrule
		bench-press & 41 & 190 & 19 & 219 & 15 & 28 & 2 & 4 & 40 & 489 & 22 & 482 \\
		front-raise & 76 & 370 & 18 & 132 & - & - & - & - & 63 & 826 & 29 & 548 \\
		push-up & 66 & 449 & 16 & 303 & 18 & 48 & 5 & 17 & 59 & 656 & 19 & 420 \\
		pull-up & 63 & 348 & 19 & 217 & - & - & - & - & 60 & 668 & 23 & 496 \\
		sit-up & 54 & 242 & 20 & 270 & - & - & - & - & 53 & 518 & 13 & 280 \\
		squat & 81 & 544 & 18 & 164 & 19 & 50 & 4 & 9 & 55 & 600 & 19 & 308 \\
        jumping jack & 49 & 350 & 26 & 713 & 17 & 49 & 5 & 20 & - & - & - & - \\
        pommel horse & 57 & 424 & 15 & 438 & 20 & 66 & 5 & 48 & - & - & - & - \\
        lunge & - & - & - & - & - & - & - & - & 33 & 441 & 20 & 294 \\
        rope push-down & - & - & - & - & - & - & - & - & 25 & 394 & 20 & 264 \\
		\midrule
		Total & \textbf{487} & 2917 & 151 & 2456 & 89 & 241 & 21 & 98 & 388 & \textbf{4501} & \textbf{165} & \textbf{3092} \\
        \midrule
        Total duration (s) &\multicolumn{2}{c}{-} & \multicolumn{2}{c}{4432} & \multicolumn{2}{c}{-} & \multicolumn{2}{c}{727} & \multicolumn{2}{c}{-} & \multicolumn{2}{c}{\textbf{4839}} \\
		\bottomrule
	\end{tabular}
\end{table*}

\subsection{Experimental Settings} \label{sec4.1}
\subsubsection{Datasets and Evaluation Metrics} \label{sec4.1.1}
\noindent To fully validate the robustness and effectiveness of our proposed GMFL-Net, we conduct extensive experiments on three challenging repetitive action counting datasets (\emph{i.e.}, RepCount-pose, UCFRep-pose, and our proposed Countix-Fitness-pose). For the RepCount-pose dataset, the training set contains 487 videos and 2,917 fine-grained pose-level annotations. Similarly, the UCFRep-pose dataset contains 110 videos, where the training set contains 89 videos and 241 fine-grained pose-level annotations.

\indent Compared to previous datasets \cite{dwibedi,levy,realworld,zhang} that typically contain only short videos, real-life scenarios often involve medium or long videos. Performing repetitive action counting in such videos is more challenging due to the presence of multiple anomalies, such as different action cycles, interruptions of repetitive actions by internal or external factors and so on. And given the limited number of publicly available datasets for the repetitive action counting task, we deliberately created a new dataset named Countix-Fitness-pose to fully validate the robustness and effectiveness of our proposed network. We select eight common fitness action classes from the original Countix \cite{dwibedi} dataset that are compatible with the pose-level method and then collect 553 videos from YouTube using the video IDs provided in Countix, ensuring coverage of different cycle lengths and complex environmental conditions. As shown in Fig.~\ref{fig5}, we add two new action classes that are not available in the other two datasets, namely lunge and rope push-down. In these two videos, there are usually some irrelevant content, such as interactions between individuals or relaxation between different actions, which constitute exceptional cases mentioned above. These instances increase the difficulty of accurate counting. And we found that these actions have a large range of variation in geometric information (\emph{i.e.}, coordinates, angles, and distances between joints) during the movements. For example, in the lunge, geometric information of the lower body is very significant for recognizing Salient pose I and II. Therefore, in our proposed MIA-Module, the module combines multi-geometric information to recognize salient poses, which results in a more detailed recognition of salient poses and effectively mitigates the impact of exceptions. By introducing these two new action classes, we not only enrich the diversity of the dataset but also enable the counting and analysis of different action focuses.

In addition, we also use the PSR mechanism proposed in the pose-level method to label our dataset. For example, for the front-raise action, the two most significant poses are arms-down and arms-raise, which are sufficient to represent the action completion. Consequently, we label each video in this dataset with the frame indexes of the two most salient poses.

Lastly, we split this new Countix-Fitness-pose dataset based on the training and testing lists provided by Countix. Detailed information about the new Countix-Fitness-pose dataset is shown in Table~\ref{table1} and Table~\ref{table2}. Compared to the dataset after applying the PSR mechanism to RepCount-pose and UCFRep-pose \cite{poserac}, our proposed dataset is richer in terms of fine-grained labelling of events. We label 4,501 fine-grained pose-level annotations in the training set and 3,092 fine-grained pose-level annotations in the test set, totalling approximately 7,593 fine-grained annotations. These richer and more specific fine-grained pose-level annotations enable the network to improve generalisation and overall performance. The training set labelled using the PSR mechanism only requires high-quality Salient pose I and II annotations without considering the duration of the videos. And the longer the duration of the test set, the more effective it is to test the performance. As shown in Table~\ref{table2}, we only consider the total duration of all the videos in the test set, which amounts to 4,839 seconds—higher than the other two datasets.

In previous research work \cite{dwibedi,zhang,hu}, two key evaluation metrics are mainly used to assess the network performance, which are Mean Absolute Error (MAE) and Off-By-One (OBO). MAE represents the average absolute error between the predictions of the model and the ground truth. On the other hand, OBO is defined as a sample count that is considered correct if the predicted value of the network does not differ from the true value by more than one (usually less than or equal to one). OBO reflects the model's fault tolerance, or its ability to be considered correct despite a certain degree of prediction error. They can be defined as follows:
\begin{equation}
	\begin{aligned}
		&MAE = \frac{1}{N} \sum_{i=1}^{N} \frac {| \widetilde{c}_i - c_i |}{\widetilde{c}_i},  \\
		&OBO = \frac{1}{N} \sum_{i=1}^{N} [| \widetilde{c}_i - c_i | \leq 1],
	\end{aligned}
	\label{eq11}
\end{equation}
where $\tilde{c}$ is the ground truth, $c_i$ is our predicted value, and $N$ is the number of videos.

\subsubsection{Training Details}

Our proposed GMFL-Net is implemented using the PyTorch-Lightning framework and trained on the NVIDIA PCle A100 GPU. When setting the initial learning rate, we can perform a training step on each small batch of data and monitor the loss change to automatically select the optimal learning rate. During training, if the loss value is not reduced for 6 consecutive epochs on the validation set, the learning rate is automatically decreased. We set Adam as the optimizer for our network.

\subsection{Ablation Studies}
In this section, we perform ablation experiments on the RepCount-pose dataset to determine the optimal configuration of each component of our GMFL-Net. These experiments examine several aspects: (1) the impact of adding different geometric information to the MIA-Module, (2) the impact of different pooling operations and regularization strategies, (3) the comparison of different global bilinear feature learning methods in the GBFL-Module, (4) the difference between the GBFL-Module and Attention mechanism used to obtain long-range dependencies, and (5) the impact of difference pooling operations in the classification head on network performance.

\begin{table}[htp]
	\centering
	\caption{Impact of different geometric information on the performance of GMFL-Net is comparatively analysed on the RepCount-pose. In this case, the additional geometric information (\emph{i.e.}, angles and distances) is added to the existing coordinate information (\textbf{\textcolor[rgb]{1,0,0}{Red}} indicates the best performance).}
	\label{table3}
	\begin{tabular}{ccccccc}
		\toprule
		the different geometric information & MAE & OBO \\
		\midrule
		Using only coordinate information & 0.243 & 0.538 \\
		Adding distance information & 0.237 & 0.553 \\
		Adding angle information & 0.246 & 0.547 \\
		Adding angle and distance information & \textbf{\textcolor[rgb]{1,0,0}{0.216}} & \textbf{\textcolor[rgb]{1,0,0}{0.586}} \\
		\bottomrule
	\end{tabular}
\end{table}

\subsubsection{Impact of Adding Different Geometric Information in GMFL-Net}
To evaluate the impact of introducing different geometric information into the MIA-Module on the network performance, we conduct experiments on the RepCount-pose dataset. As shown in Table~\ref{table3}, the best performance is achieved when two types of geometric information (\emph{i.e.}, angle and distance between joints) are added to the MIA-Module. In addition, adding either angle or distance information to the existing coordinate information improves performance as well. This is because the angles and distances between joints change significantly during the movement, which can effectively reflect the movement trend and action details. The changes in geometric information for different actions are unique and specific. Therefore, the combined effect of coordinate information, angles and distances between joints has a positive effect on improving network performance.
\begin{table}[htp]
	\centering
	\caption{Impact of different pooling operations and regularization strategies on the GBFL-Module, where "$*$" denotes element-by-element product, "$+$" denotes summation, and "$-$" denotes subtraction. Operations (1) and (2) represent the pooling operation in Equations (8) and (9), respectively. Regularization is the operation of Equation (12).}
	\label{table4}
	\begin{tabular}{ccccccc}
        \toprule
        Operation & Operation & Regulari- & \multirow{2}{*}{MAE} & \multirow{2}{*}{OBO} \\
        (1) & (2) & zation \\
        \midrule
        max-pooling & max-pooling & * & 0.252 & 0.507 \\
        max-pooling & max-pooling & + & 0.243 & 0.513 \\
        max-pooling & max-pooling & - & 0.236 & 0.531 \\
        avg-pooling & avg-pooling & * & 0.263 & 0.553 \\
        avg-pooling & avg-pooling & + & 0.247 & 0.528 \\
        avg-pooling & avg-pooling & - & \textbf{\textcolor[rgb]{1,0,0}{0.216}} & \textbf{\textcolor[rgb]{1,0,0}{0.586}} \\
        \bottomrule
    \end{tabular}
\end{table}

\subsubsection{Impact of Different Pooling Operations and Regularization Strategies in GBFL-Module}
Apart from avg-pooling, max-pooling can also be used to extract salient features from the geometric feature map $F_L$ learned from the MIA-Module. To investigate the optimal combination of global feature fusion, we conduct experiments using different pooling operations and regularization strategies. Specifically, the different pooling operations include avg-pooling and max-pooling, while the regularization strategies include element-by-element dot product, summation, and subtraction. As shown in Table~\ref{table4}, we use the same pooling operations for Equation (8) and Equation (9), which allows the network to focus more on similar types of features. The use of different pooling operations can lead to varying feature representations, causing them to interfere with each other. As can be seen, our method performs best when both Equation (8) and Equation (9) use avg-pooling and the regularization strategy of Equation (12) is subtraction. This is because avg-pooling smoothes the feature map and captures the overall trend of local features. The subtraction operation, on the other hand, minimizes the negative influence of the absolute position of the local feature map $F_L$ on the global feature $F_G$. Therefore, this combination is most effective.

\begin{table}[htp]
	\centering
	\caption{Impact of different global bilinear feature learning methods on the GBFL-Module, where Product is element by element product.}
	\label{table5}
	\begin{tabular}{ccccccc}
        \toprule
        Learning  & Formulation & \multirow{2}{*}{MAE} & \multirow{2}{*}{OBO} \\
           methods     &  in Equation(10)  \\
        \midrule
        Summation & $G_C + G_N$ & 0.276 & 0.513 \\
        Product & $G_C \cdot G_N$ & 0.226 & 0.517 \\
        Grand Mean & $(G_C + G_N)/2$ & 0.268 & 0.521 \\
        Quadratic Mean & $\sqrt{(G_C^2 + G_N^2)}$ & 0.233 & 0.557  \\
        Harmonic Mean & $2G_C G_N/(G_C + G_N)$ & 0.235 & 0.568  \\
        Geometric Mean & $\sqrt{G_C \cdot G_N}$ & \textbf{\textcolor[rgb]{1,0,0}{0.216}} & \textbf{\textcolor[rgb]{1,0,0}{0.586}} \\
        \bottomrule
    \end{tabular}
\end{table}

\subsubsection{Impact of Different Global Bilinear Feature Learning Methods in GBFL-Module}
In order to further investigate the global feature aggregation methods based on the point-wise global feature $G_N$ and channel-wise global feature $G_C$, we explore six different aggregation methods in Table~\ref{table5}, which are summation, element-by-element product, grand mean, quadratic mean, harmonic mean, and geometric mean. The experimental results show that the quadratic mean, harmonic mean, and geometric mean are more effective in estimating the global information in spatial and channel dimensions, as they allow for a more comprehensive and robust treatment and representation of the data features, avoiding the limitations and numerical problems associated with simple summation, element-by-element product, and grand mean. Notably, compared to other means, the use of geometric means for global feature learning proves to be the most effective.
\begin{table}[htp]
	\centering
	\caption{Comparison of differences between GBFL-Mdoule and different attention mechanism in terms of Params, FLOPs, and evaluation metrics.}
	\label{table6}
	\begin{tabular}{ccccccc}
        \toprule
        \multirow{2}{*}{Methods} & Params & FLOPs & \multirow{2}{*}{MAE} & \multirow{2}{*}{OBO} \\
        &  ($\times10^3$) & ($\times10^7$)  \\
        \midrule
        Self Attention & 49.28 & 10.40 & 0.226 & 0.567 \\
        Offset Attention & 41.47 & 8.78 & 0.237 & 0.556 \\
        Point Transformer & 25.48 & 1391.67 & 0.236 & 0.547 \\
        GBFL-Module & \textbf{\textcolor[rgb]{1,0,0}{6.56}} & \textbf{\textcolor[rgb]{1,0,0}{1.41}} & \textbf{\textcolor[rgb]{1,0,0}{0.216}} & \textbf{\textcolor[rgb]{1,0,0}{0.586}} \\
        \bottomrule
    \end{tabular}
\end{table}

\subsubsection{Comparison of Difference Between the GBFL-Mdoule and Attention Mechanism}
It is well known that the attention mechanism is one of the best available methods for global feature aggregation, and our proposed GBFL-Module is also designed for global feature learning. Therefore, we compare the GBFL-Module with the self-attention mechanism, the offset attention mechanism, and the Transformer block in Point Transformer in terms of the number of Params, FLOPs, and evaluation metrics. As shown in Table~\ref{table6}, the GBFL-Module achieves the best performance with the lowest computational cost. In contrast, other methods may excessively rely on point-wise or channel-wise information, leading to the generation of redundant feature representations, which in turn affects network performance.
\begin{table}[htp]
	\centering
	\caption{Impact of Difference Pooling Operations in Equation (13) in The Classification Head on Model Performance.}
	\label{table7}
	\begin{tabular}{ccccccc}
        \toprule
        Operation in Equation(13)& MAE & OBO \\
        \midrule
        max-pooling + max-pooling  & 0.242 & 0.557 \\
        avg-pooling + avg-pooling  & 0.236 & 0.563 \\
        avg-pooling + max-pooling  & \textbf{\textcolor[rgb]{1,0,0}{0.216}} & \textbf{\textcolor[rgb]{1,0,0}{0.586}} \\
        \bottomrule
    \end{tabular}
\end{table}

\subsubsection{Impact of Difference Pooling Operations in Equation (13)}
To explore the best way to generate the optimal feature representation in the Classification Head, we conduct experiments to investigate the impact of combining different pooling operations in the Classification Head. In contrast to the ablation study in Table~\ref{table4}, which uses the same pooling operations to maintain feature consistency and coherence when constructing global features, this subsection combines different types of pooling operations to extract rich and discriminative features. The generated diverse feature sets for the Classification Head can increase the network's sensitivity to subtle features. As shown in Table~\ref{table7}, we observe that the proposed method performs best when the output of avg-pooling and max-pooling are concatenated. This is due to max-pooling highlights the most salient features in the global bilinear feature, while avg-pooling captures the overall trend in the global bilinear feature. By concatenating them, we can effectively capture features at different types, reduce the loss of useful information and significantly improve the expressiveness of the network.

\begin{table*}[t]
	\centering
	\caption{Performance of our proposed method on RepCount-pose, UCFRep-pose, and Countix-Fitness-pose. RepCount and RepCount-pose (as well as UCFRep and UCFRep-pose, and Countix-Fitness-pose) share the same training and test sets, and the fine-grained annotations are the pose-level.}
	\label{table8}
	\begin{tabular}{cccccccccc}
		\toprule
		\multirow{2}{*}{Category} & \multirow{2}{*}{Methods} & \multicolumn{2}{c}{RepCount-pose} & \multicolumn{2}{c}{UCFRep-pose} & \multicolumn{2}{c}{Countix-Fitness-pose} \\
		\cmidrule(lr){3-4} \cmidrule(lr){5-6} \cmidrule(lr){7-8}
		&& MAE $\downarrow$ & OBO $\uparrow$ & MAE $\downarrow$ & OBO $\uparrow$ & MAE $\downarrow$ & OBO $\uparrow$ \\
		\midrule
		\multirow{7}{*}{\shortstack[l]{video-level}} & RepNet \cite{dwibedi} & 0.995 & 0.013 & 0.981 & 0.018 & 0.432 & 0.393 \\
		& X3D \cite{x3d} & 0.911 & 0.106 & 0.982 & 0.331 & 0.956 & 0.126  \\
		& Zhang et al. \cite{zhang} & 0.879 & 0.155 & 0.762 & 0.412 & 0.457 & 0.377  \\
		& TANet \cite{liu} & 0.662 & 0.099 & 0.892 & 0.129 & 0.507 & 0.369  \\
		& Video Swin Transformer \cite{swin} & 0.576 & 0.132 & 1.122 & 0.033 & 0.706 & 0.205  \\
		& Huang et al. \cite{huang} & 0.527 & 0.159 & 1.035 & 0.015 & 1.029 & 0.041 \\
		& TransRAC \cite{hu} & 0.443 & 0.291 & 0.581 & 0.329 & 0.478 & 0.283 \\
		\midrule
		\multirow{2}{*}{\shortstack[l]{pose-level}} & PoseRAC \cite{poserac} & 0.236 & 0.560 & 0.312 & 0.452 & 0.387 & 0.497 \\
		& GMFL-Net(ours) &  \textbf{\textcolor[rgb]{1,0,0}{0.216}} &  \textbf{\textcolor[rgb]{1,0,0}{0.586}} &  \textbf{\textcolor[rgb]{1,0,0}{0.259}} &  \textbf{\textcolor[rgb]{1,0,0}{0.650}} & \textbf{\textcolor[rgb]{1,0,0}{0.269}}  &  \textbf{\textcolor[rgb]{1,0,0}{0.594}} \\
		\bottomrule
	\end{tabular}
\end{table*}
\begin{figure*}[htp]  
	\centering
	\centering	
	\includegraphics[width=18cm,scale=1]{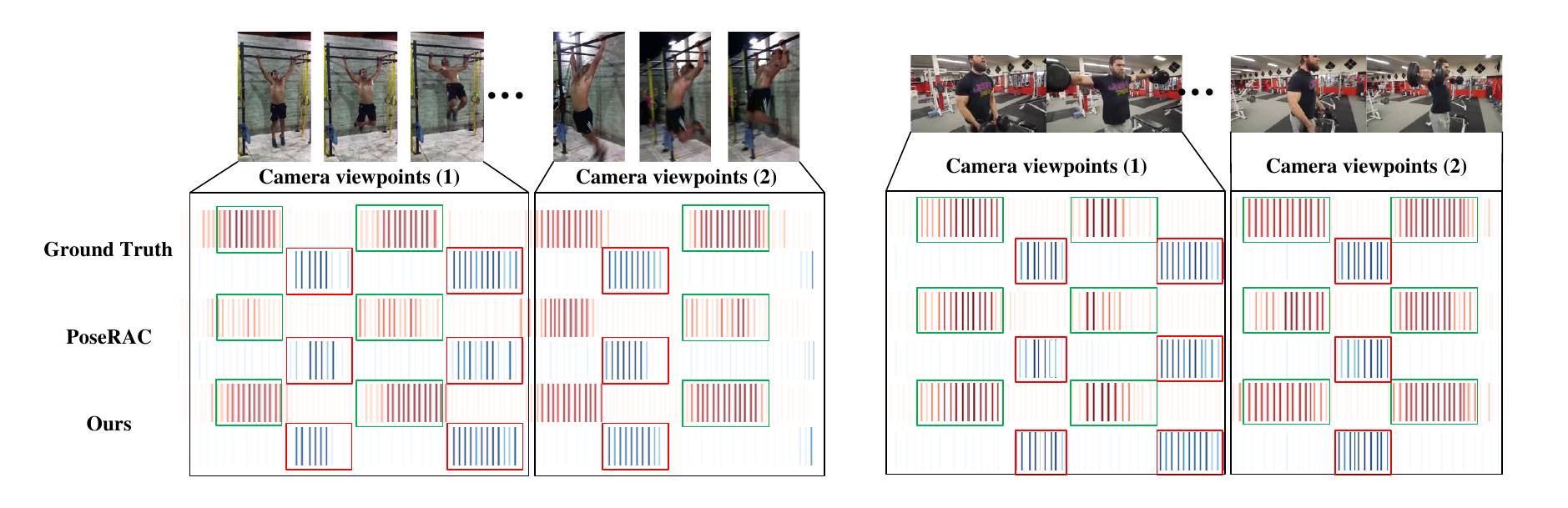}
	\caption{Visualisation of the pose mapping. This visualisation shows the comparison between our method and PoseRAC as well as the ground truth. A darker red colour in the graph means that it is more likely to represent Salient pose I, while a darker blue colour means that it is more likely to represent Salient pose II.}
	\label{fig6}
\end{figure*}
\subsection{Benchmark Comparison}
\subsubsection{RepCount-pose}
\noindent As shown in Table~\ref{table8}, our proposed GMFL-Net is compared with some previous state-of-the-art methods for repetitive action counting on the RepCount-pose dataset \cite{dwibedi,x3d,zhang,liu,swin,huang,hu,poserac}. It can be observed that our proposed network outperforms all the previous methods in two key evaluation metrics, \emph{i.e.}, MAE of 0.216 and OBO of 0.586. Because our pose-level method focuses on the most central human poses in repetitive actions, it eliminates the interference of redundant information such as background details in video-level methods. Compared to the state-of-the-art video-level method TransRAC \cite{hu}, our method reduces MAE by 22.7\% and improves OBO by 29.5\%. In addition, our method introduces multi-geometric information on top of a single coordinate information, which improves the characterisation of different action details. While previous pose-level methods ignore the importance of local detail features of the action from a global perspective, our method effectively improves the network performance by learning the inter-dependence of local and global features in the geometric information. As a result, our method reduces MAE by 2.0\% and improves OBO by 2.6\%.
\subsubsection{UCFRep-pose}
\noindent As shown in Table~\ref{table8}, our proposed GMFL-Net is compared with some state-of-the-art methods on the UCFRep-pose dataset \cite{dwibedi,x3d,zhang,liu,swin,huang,hu,poserac}. Similarly, the performance of our method on this dataset is quite excellent, MAE is 0.259 and OBO is 0.650. Compared to the pose-level method \cite{poserac}, our method achieves a reduction of 5.3\% in MAE and an improvement of 19.8\% in OBO. On the other hand, compared to the state-of-the-art video-level method \cite{hu}, our method achieves a reduction of 32.2\% in MAE and an improvement of 32.1\% in OBO. This highlights the improvement in network effectiveness by introducing geometric information and learning the dependencies between its local and global features.

\subsubsection{Countix-Fitness-pose}
\noindent To ensure a fair and unbiased evaluation, we adapt the output layers of previous methods \cite{dwibedi,x3d,zhang,liu,swin,huang,hu,poserac} to meet the requirements of our proposed dataset and conduct extensive experiments on this dataset. As shown in Table~\ref{table8}, we can observe that our proposed GMFL-Net achieves MAE of 0.269 and OBO of 0.594 on the Countix-Fitness-pose dataset, maintaining a leading position. Compared to the pose-level method \cite{poserac}, our method reduces MAE by 11.8\% and improves OBO by 9.7\%. We can also observe that our method is far ahead of the video-level method \cite{dwibedi,x3d,zhang,liu,swin,huang,hu} across the board. Despite the fact that the action classes in this dataset are more diverse and the test set duration is longer, there are also some influences such as changes in camera viewpoint and exceptions in the video, our method is still stable in recognizing salient poses and reduces the number of misdetections with the introduction of multi-geometric information. In addition, by mining and learning the inter-dependence between local and global features in the multi-geometric information, we are able to capture the motion details between different actions, which leads to more accurate recognition of salient poses. As a result, our method makes the network more robust and efficient.

\subsection{Qualitative Evaluation}
\indent To validate the effectiveness of our method, we visualise the output of the pose mapping in Fig.~\ref{fig6}. We found that because PoseRAC relies only on a single coordinate information, it is susceptible to the problem of viewpoint difference when the camera viewpoint changes, resulting in the network failing to recognize the change of salient pose in time. In contrast, our method introduces multi-geometric information, enhances the capture of the overall motion trend and details of the action, and appropriately improves the interaction between global and local features, which improves the stability of the salient pose recognition and achieves satisfactory results in terms of performance.

\section{Conclusion} \label{sec5}
In this paper, we propose a simple and effective GMFL-Net for repetitive action counting. Based on the pose-level methods, we introduce multi-geometric information hidden between joints, including features such as coordinates, angles, and distances. This geometric information reveals the overall motion trend and details of the action, which helps to improve the recognition accuracy of repetitive actions. Specifically, our method introduces the innovative MIA-Module and GBFL-Module, where the MIA-Module improves the information representation by fusing multi-geometric features, while the GBFL-Module further enhances the feature representation through the point-wise and channel-wise global feature learning. The results of ablation studies indicate that the introduction of multi-geometric information and global feature learning is crucial for performance improvement. In addition, given the limited diversity of existing pose-level datasets, we construct a new dataset, Countix-Fitness-pose, which adds two new action classes not covered in other datasets. The dataset contains richer fine-grained annotations and the test set has longer duration, thus providing a more challenging and richer data resource for future research. Finally, we conduct extensive experiments on three challenging datasets (RepCount-pose, UCFRep-pose, and Countix-Fitness-pose). The results from ablation experiments, qualitative evaluations, and comparative experiments comprehensively demonstrate the effectiveness of our proposed method.

\vfill

\end{document}